\documentclass[letterpaper]{article} 
\usepackage{aaai25}  
\usepackage{times}  
\usepackage{helvet}  
\usepackage{courier}  
\usepackage[hyphens]{url}  
\usepackage{graphicx} 
\usepackage{natbib}  
\usepackage{caption} 
\usepackage{algorithm}
\usepackage{algorithmic}
\usepackage{amsmath}
\usepackage{amssymb}
\usepackage{multirow}

\usepackage{newfloat}
\usepackage{listings}
\DeclareCaptionStyle{ruled}{labelfont=normalfont,labelsep=colon,strut=off} 
\floatstyle{ruled}
\newfloat{listing}{tb}{lst}{}
\floatname{listing}{Listing}
\title{Noise-Injected Spiking Graph Convolution for Energy-Efficient 3D Point Cloud Denoising}
\author{
    Zikuan Li\textsuperscript{\rm 1}, 
    Qiaoyun Wu\textsuperscript{\rm 3}\thanks{The corresponding authors.}, 
    Jialin Zhang\textsuperscript{\rm 2}, 
    Kaijun Zhang\textsuperscript{\rm 2}, 
    Jun Wang\textsuperscript{\rm 1,\rm 2} 
}
\affiliations{
    \textsuperscript{\rm 1}School of Computer Science and Technology, Nanjing University of Aeronautics and Astronautics\\
    \textsuperscript{\rm 2}College of Mechanical and Electrical Engineering, Nanjing University of Aeronautics and Astronautics\\
    \textsuperscript{\rm 3}School of Artificial Intelligence, Anhui University\\
    lzkuan@nuaa.edu.cn, wuqiaoyun@ahu.edu.cn, wjun@nuaa.edu.cn

}

\usepackage{bibentry}

\begin{document}

\maketitle

\begin{abstract}
Spiking neural networks (SNNs), inspired by the spiking computation paradigm of the biological neural systems, have exhibited superior energy efficiency in 2D classification tasks over traditional artificial neural networks (ANNs). However, the regression potential of SNNs has not been well explored, especially in 3D point cloud processing.
In this paper, we propose noise-injected spiking graph convolutional networks to leverage the full regression potential of SNNs in 3D point cloud denoising. 
Specifically, we first emulate the noise-injected neuronal dynamics to build noise-injected spiking neurons.
On this basis, we design noise-injected spiking graph convolution for promoting disturbance-aware spiking representation learning on 3D points.
Starting from the spiking graph convolution, we build two SNN-based denoising networks.
One is a purely spiking graph convolutional network, which achieves low accuracy loss compared with some ANN-based alternatives, while resulting in significantly reduced energy consumption on two benchmark datasets, PU-Net and PC-Net.
The other is a hybrid architecture that combines ANN-based learning with a high performance-efficiency trade-off in just a few time steps.
Our work lights up SNN’s potential for 3D point cloud denoising, 
injecting new perspectives of exploring the deployment on neuromorphic chips while paving the way for developing energy-efficient 3D data acquisition devices.
\end{abstract}

\begin{links}  
\link{Code}{https://github.com/Miraclelzk/NI-SGCN}  
\end{links}

\section{Introduction}
\label{sec:intro}
Intelligent robotic systems are often equipped with a 3D data acquisition device,
which provides detailed point clouds for environment perception.
However, raw point clouds are often corrupted by noise,
stemming from sensor imperfections, environmental interference, and the probabilistic nature of data capture. 
Point cloud denoising endeavors to extricate coherent geometrical information from raw scans, thereby enhancing the precision of the data and fortifying its applicability in advanced perception tasks.
With the continuous advancement of deep learning, learning-based point cloud denoising methods have achieved remarkable progress~\cite{rakotosaona2020pointcleannet, zhang2020pointfilter, luo2021score}. 
However, these ANN-based methods typically demand numerous multiply-accumulate operations, leading to high computational costs and energy consumption.
This poses significant challenges when deploying such methods on 3D scanning devices, which have resources and battery constraints.

Spiking Neural Networks (SNNs), using spike-driven communication, have emerged as a prominent solution to mitigate the exiguity of energy efficiency inherent in contemporary deep learning.
In SNNs, all information is encoded within spiking signals rather than floating-point representations, 
which allows SNNs to adopt spike-based accumulate (AC) operations instead of energy-hungry multiply-accumulate (MAC) operations, thus leading to extremely low energy consumption~\cite{kim2020spiking}.
The expeditious progression in the domain of neuromorphic engineering has culminated in the advent of sophisticated chips like Loihi~\cite{davies2018loihi}, which significantly augment the energy efficiency inherent to SNNs. This advancement brings us to the future of integrating intelligent neuromorphic processors into the fabric of quotidian existence.

Despite SNNs have been successfully applied in the field of neuromorphic computing~\cite{cao2024spiking}, the current body of research on SNNs has been predominantly directed towards classification endeavors, inadvertently neglecting the exploration of their regression competencies, notably within the domain of 3D point cloud denoising. 
Hence, to develop a SNN-based regression algorithm supporting effective point cloud denoising whilst maintaining a profile of low energy expenditure, is imperative. 
However, two key concerns deserve much attention in extending SNNs to the domain: 
(1) stochastic noise will bring spiking disturbance to deterministic models, directly undermining their robustness, and (2) semantically similar structures within a point cloud can facilitate the mutual perception of perturbations, thereby enhancing the discriminative capacity of the system to detect and respond to noises in the data.

This paper explores SNNs for energy-efficient 3D point cloud denoising.
For the first concern, we draw inspiration from the inherent nondeterministic and noisy nature of neural computations, and build noise-injected spiking neurons to yield flexible and and reliable learning on 3D points. 
For the second concern, we borrow concepts from EdgeConv~\cite{wang2019dynamic}, especially learning to capture semantically similar structures by dynamically updating a graph of point relationships.
We integrate our spiking neurons into the convolution and design the noise-injected spiking graph convolution, which increases the representation power for discerning disturbances while efficiently propagating information via sparse spiking signals. 
On this basis, we propose two variants of noise-injected spiking graph convolutional networks (NI-SGCN) for the denoising of 3D point cloud data.
The first model is articulated in its entirety within the spiking paradigm, denoted as NI-PSGCN, exemplifying the Purest form of spiking computation.
The second model, NI-HSGCN, represents a Hybrid construct that integrates select learning operations from artificial neural networks (ANNs), thereby harnessing the complementary strengths of both SNNs and ANNs within a cohesive architecture. The main contributions are as follows:
\begin{itemize}
    \item We define the noise-injected spiking neuron for nondeterministic spiking learning on 3D points. We show the neuron leads to SNNs with competitive performance and improved robustness when facing challenging disturbances compared with deterministic spiking neurons.
    \item We design the noise-injected spiking graph convolution, capable of exploiting semantically similar structures to facilitate spiking representation learning while also enhancing the information flow efficiency.  
    \item We develop two noise-injected spiking graph convolutional networks for 3D point cloud denoising, which significantly reduces energy consumption. To the best of our knowledge, our work is the first to employ spiking neural networks for energy-efficient point cloud denoising, while maintaining high accuracy on the PU-Net dataset~\cite{yu2018pu} and PC-Net dataset~\cite{rakotosaona2020pointcleannet}.
\end{itemize}

\section{Related Work}
\subsection{Denoise on Point Clouds}
Most learning-based point cloud denoising methods evolve from foundational point cloud processing techniques~\cite{preiner2014continuous}. 
The development of PointNet and PointNet++~\cite{qi2017pointnet++} enables the direct convolution of point sets, paving the way for more advanced approaches. 
Building on these advancements, Wang $et$ $al.$~\cite{wang2019dynamic} introduce a graph convolutional architecture that uses nearest-neighbor graphs derived from point sets to generate rich feature representations.

PointCleanNet (PCN)~\cite{rakotosaona2020pointcleannet} employs a two-level network to first remove outlier points and then learn the motion coordinates of the noisy point cloud, transforming it into a cleaner version. 
Pointfilter~\cite{zhang2020pointfilter} uses clean normals as a supervisory signal to analyze the model's latent surface information, effectively removing noise while preserving the sharp edges of the point cloud. 

Luo $et$ $al.$ introduce ScoreDenoise, a score-based denoising method that models the gradient log of the noise-convolved probability distribution for point cloud patches~\cite{luo2021score}. 
Chen $et$ $al.$\cite{chen2019multi} propose a multi-block denoising approach based on low-rank matrix recovery with graph constraints and later developed RePCD~\cite{chen2022repcd}, a feature-aware recurrent network.
Edirimuni $et$ $al.$~\cite{de2023iterativepfn} criticize RePCD for its lack of iterative noise reduction during testing and propose IterativePFN, an iterative point cloud filtering network that explicitly models the iterative filtering process internally. 
Wei $et$ $al.$\cite{wei2024pathnet} propose PathNet, a path-selective point cloud denoising framework that adapts its approach based on varying levels of noise and the distinct geometric structures of the points.

\subsection{Spiking Neural Networks}
SNNs are regarded as the third generation of neural networks, inspired by brain-like computing processes that use event-driven signals to update neuronal nodes~\cite{cao2024spiking}. 
Unlike conventional ANNs, spiking neurons operate on discrete-time events rather than continuous values, making SNNs more energy and memory-efficient on embedded platforms~\cite{wu2019direct}.

One significant challenge with SNNs is the effective training and optimization of network parameters.
Currently, there are two primary methods for developing deep SNN models: ANN-to-SNN conversion and direct training. 
In ANN-to-SNN conversion, ReLU activation layers are replaced with spiking neurons to replicate the behavior of the original ANN.
However, these converte SNNs often require substantial inference time and memory, resulting in increased latency and decreased energy efficiency, which undermines the advantages of spiking models~\cite{roy2019towards}. 
In contrast, direct training involves designing surrogate gradients for backpropagation or using gradients with respect to membrane potentials to train SNNs from scratch. 
Models trained directly tend to reduce spiking time latency and are more suitable for practical applications. 
However, for large-scale tasks, they often do not match the accuracy of conversion-based approaches or ANNs~\cite{roy2019towards}.

To enhance SNN performance and bridge the gap between ANNs and SNNs, several advancements have been made. 
Wu $et$ $al.$~\cite{wu2019direct} introduce neuron normalization to balance firing rates and preserve important information.
The QIF neuron~\cite{brunel2003firing} simulates neuronal electrical activity by extending the standard Integrate-and-Fire (IF) neuron with a quadratic nonlinearity, offering a more accurate representation of the neuron's membrane potential.
The KLIF neuron~\cite{jiang2023klif} is a novel k-based leaky integrate-and-fire (LIF) neuron designed to enhance the learning capabilities of spiking neural networks.
Spiking neurons with noise-injected dynamics are considered more biologically realistic. 
Rao $et$ $al.$~\cite{rao2004bayesian} develope small noise-spiking neural networks to perform probabilistic reasoning, effectively improving network robustness. 
However, integrating these methods into arbitrary network architectures remains challenging.

\subsection{Spiking Neural Networks on Point Cloud}
Recent research efforts are exploring the application of SNNs in point cloud processing. 
Lan $et$ $al.$~\cite{lan2023efficient} propose an efficient unified ANN-SNN conversion method for point cloud and image classification, significantly reducing time steps for a fast and lossless transformation. 
Ren $et$ $al.$~\cite{ren2024spiking} extend PointNet to SNNs and develop Spiking PointNet, while Wu $et$ $al.$~\cite{wu2024point} introduce a point-to-spike residual learning network for point cloud classification. 
Despite these advances, there are relatively few studies combining SNNs with point cloud data, and most focus on classification tasks. 
To our knowledge, we are the first to apply SNNs to point cloud denoising.

\section{Methods}
We propose noise-injected spiking graph convolutional networks for 3D point cloud denoising striking a good balance between effectiveness and efficiency.
We first define noise-injected spiking neurons, which take advantage of non-deterministic, noisy neurodynamic computations derived from the brain to enhance computational robustness.
We then integrate noise-injected spiking neurons into traditional graph convolution~\cite{wang2019dynamic} to create noise-injected spiking graph convolution, enhancing 3D point feature learning from spike sequences.
Finally, we build 3D spiking denoising networks on the basis of the proposed spiking graph convolution.

\subsection{Noise-Injected Spiking Neurons}
Inspired by the inherent nondeterministic characteristics of neural computations~\cite{ma2023exploiting}, 
we define the noise-injected spiking neuron for spike computation.

\paragraph{Integrate-and-Fire spiking neurons.}
Bio-inspired spiking neurons are designed to mimic the actual signal processing in the brain.
In this paper, we employ the simplest model of spiking neurons, the Integrate-and-Fire model, which is defined as:

\begin{equation}
\label{eq:men_v}
U_t=V_{t-1}+I_t
\end{equation}
\begin{equation}
\label{eq:spike}
S_t=\Theta(U_t-V_{th})
\end{equation}
\begin{equation}
\label{eq:after}
V_t=U_t(1-S_t)+V_{reset}S_t
\end{equation}
Where $U_t$ denotes the membrane potential at time step $t$. 
$S_t$ is the spike output, which occurs when $U_t$ exceeds a threshold $V_{th}$.
$I_t$ is the input current at time step $t$.
$\Theta(\cdot)$ is the Heaviside step function, and $V_t$ is the membrane potential after a spike is triggered. 
We employ a “hard reset” method~\cite{fang2021incorporating} in Eq.~(\ref{eq:after}), meaning that after a spike ($S_t = 1$), the membrane potential $V_t$ resets to $V_{reset} = 0$.

\paragraph{Noise-injected Integrate-and-Fire spiking neurons.}  
To improve the robustness against challenging disturbances, we inject noise into the IF spiking neuron.
Specifically, we add a Gaussian noise term to Eq.~(\ref{eq:men_v}).
The dynamics of the noise-injected IF spiking neuron (NIIF) are defined as: 
\begin{equation}
\label{eq:noise}
U_t=V_{t-1}+I_t+\epsilon
\end{equation}
Here, the noise term $\epsilon$ is drawn from a Gaussian distribution, ${\epsilon}{\thicksim}N({\mu},{\sigma}^2)$, where ${\mu}$ is the mean and ${\sigma}$ is the standard deviation of the distribution.
Additionally, the firing and resetting dynamics of noisy spiking neurons remain unchanged, namely Eq.~(\ref{eq:spike})(\ref{eq:after}).

\begin{figure}
\begin{center}
\includegraphics[width=1\linewidth]{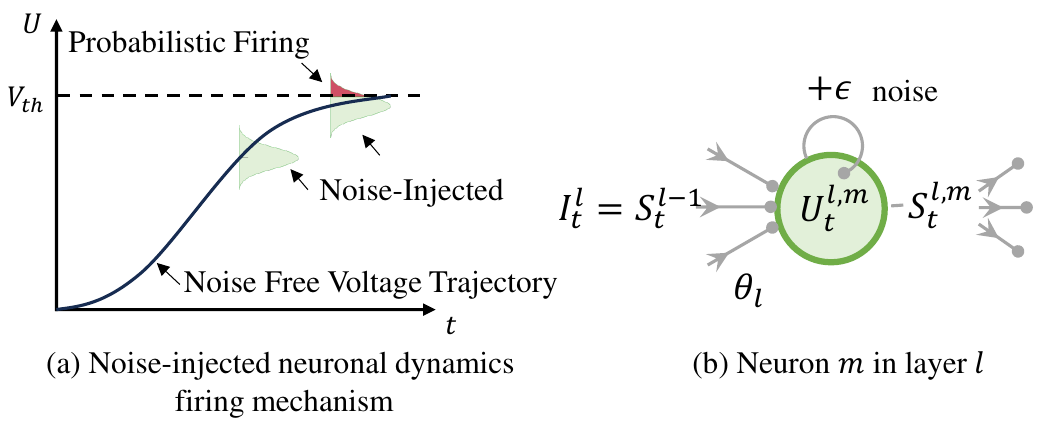}
\end{center}
   \caption{The probabilistic firing mechanism and calculation process of the NIIF neuron.}
\label{fig:noise}
\end{figure}

Fig.~\ref{fig:noise} illustrates the probabilistic firing mechanism and calculation process of the NIIF neuron. 
The firing probability is represented by the membrane noise cumulative distribution function, depicted by the shaded red area under the noisy voltage distribution.
At the time step $t$, the membrane potential $U^{l,m}_t$ and the spike output $S^{l,m}_t$ of the $m$-th neuron in the $l$-th layer, become random variables due to the injected noise~\cite{ma2023exploiting}.
The computation for updating synaptic weights denotes as $\theta_l$.
With the noise, we obtain the firing probability distribution of NIIF based on the threshold firing mechanism:
\begin{equation}
P[S_t=1]=F_{\epsilon}(U_t-V_{th}),
\end{equation}
where $S_t$ is the spike state, and $F_{\epsilon}$ is the cumulative distribution function of the noise.
The difference $U_t-V_{th}$ governs the firing probability. 
Specifically, it relates to previous literature on escape noise models~\cite{jolivet2006predicting}.

\subsection{Noise-Injected Spiking Graph Convolution}
Inspired by the discriminative representation capability of the dynamic graph convolution from EdgeConv~\cite{wang2019dynamic}, we combine the noise-injected spiking neuron with the graph convolution to promote disturbance-aware spiking representation learning on 3D points.

\paragraph{Graph convolution.}
Graph convolution~\cite{wang2019dynamic} is defined to learn the local geometric structures by iteratively computing edge features between points.
For a point cloud $X = \{x_1, x_2, \ldots, x_n\}$, where each $x_i \in \mathbb{R}^3$, we first construct a graph $G = (X, E)$, where $X$ is the vertex set and the edge set is constucted based on the $k$-nearest neighbor relationship.

At the $l$-th graph convolutional layer, the edge feature $e^l_{ij}$ between a feature $f_i$ and its neighbor feature $f_j$ is defined as:
\begin{equation}
e^l_{ij} = h_\theta(f^{l-1}_i, f^{l-1}_j - f^{l-1}_i)
\label{eq:edgeconv}
\end{equation}
Here, $h_\theta$ is a parametric function, typically modeled as a fully connected layer, which processes the concatenation of the feature $f^{l-1}_i$ from $x_i$ and the relative feature displacement $f^{l-1}_j - f^{l-1}_i$. 
This helps capture local neighborhood information and directional relationships between points. 
The edge features are then aggregated for each feature $f_i$ using a symmetric function such as max pooling:
\begin{equation}
f^{l}_i= \max_{j \in \mathcal{N}(i)} e^l_{ij}
\label{eq:max}
\end{equation}
where $\mathcal{N}(i)$ denotes the neighbors of feature $f_i$. 
The aggregation is invariant to the order of neighbors and extracts robust local features.
By stacking multiple graph convolution layers, the network refines point features iteratively to capture increasingly complex geometric patterns and spatial hierarchies within the point cloud.

\begin{figure}
\begin{center}
\includegraphics[width=0.95\linewidth]{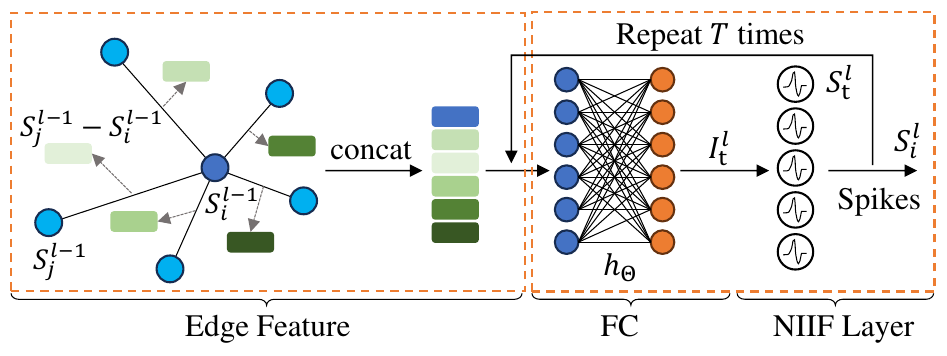}
\end{center}
   \caption{An illustration of noise-injected spiking graph convolution.}
\label{fig:SEdgeConv}
\end{figure}

\paragraph{Noise-injected spiking graph convolution (NI-SGC).}
To leverage graph convolution for capturing local and global geometric features in point clouds while enhancing the dynamics and expressiveness of the network with spiking neurons, we combine the NIIF neuron with the graph convolution to design the noise-injected spiking graph convolution as:
\begin{equation}
\begin{aligned}  
I_t^l(x_i)&=\{h_\theta(S^{l-1}_{t}(x_i), S^{l-1}_{t}(x_j) - S^{l-1}_{t}(x_i))\}_{j \in \mathcal{N}(i)}, \\ 
U_t^l(x_i)&=V_{t-1}^l(x_i)+I_t^l(x_i)+\epsilon, \\  
S_t^l(x_i)& =\Theta\left(U_t^l(x_i)-V_{t h}\right) ,\\  
V_t^l(x_i)&=U_t^l(x_i)\left(1-S_t^l(x_i)\right)+V_{\text {reset }} S_t^l(x_i),
\end{aligned}
\label{eq:update}
\end{equation}
Where $I_t^l(x)$ is the input current inspired by Eq.~(\ref{eq:edgeconv}) and~(\ref{eq:max}).
The differences lie in two aspects.
The first is the spiking feature inputs from the $(l-1)$-th layer.
Consequently, the computation of $I_t^l(x)$ can be simplified to AC operations, with weight accumulation occurring only when neighboring neurons generate spikes in the $(l-1)$-th layer.
$U^l_t(x)$ denotes the membrane potential of the neuron at 3D position $x$ in the $l$-th layer at time step $t$.
$V_{t-1}^l$ represents the membrane potential at time $(t - 1)$.
The Heaviside step function $\Theta(\cdot)$ is used for spiking determination.
Fig.~\ref{fig:SEdgeConv} illustrates the dynamics of the proposed spiking graph convolution neuron.
At the $l$-th layer, the neuron receives spike inputs from the neighborhood $\mathcal{N}_{i}$ of the $(l - 1)$-th layer. 
NI-SGC then constructs the edge feature for the spikes $S^{l-1}$ from the $(l - 1)$-th layer, followed by processing through FC and NIIF Layers.

\begin{figure}
\begin{center}
\includegraphics[width=1\linewidth]{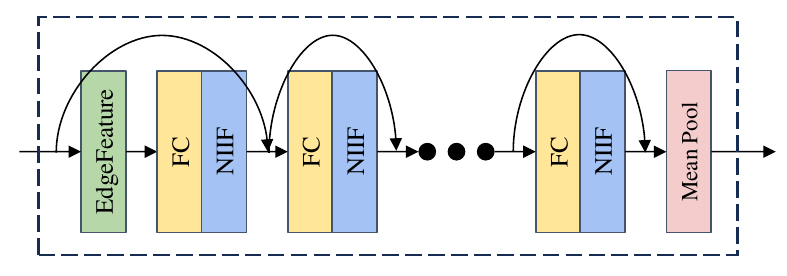}
\end{center}
   \caption{An illustration of dense NI-SGC block.}
\label{fig:NISDenseEdgeConv}
\end{figure}

\paragraph{Dense NI-SGC block.}
\label{subsec:block}
The spiking graph convolution (NI-SGC) can efficiently extract multi-scale and non-local features for each point, while dense connections can provide richer contextual information, which is much suitable for denoising tasks~\cite{liu2019densepoint}. 
Hence, we design dense NI-SGC block for point cloud denoising by elaborately combining NI-SGC, FC, and NIIF. The $r$-th NI-SGC block can be formulated as:
\begin{equation}
S_t^{r+1}(x_i)=\operatorname{mean}_{j \in \mathcal{N}(i)} H_\theta (S_t^{r}(x_i), S_t^{r}(x_j)-S_t^{r}(x_i) )
\end{equation}
where $S_t^{r}=\left\{S_t^{r}(x_i)\right\}_{i=1}^N$ are spiking feature representations in a high-dimensional space, serving as the input to the $r$-th block. 
$H_\theta$ represents a densely connected FC parameterized by $\theta$, $\mathcal{N}(i)$ denotes the neighborhood of spiking feature $S_t(x_i)$.
The other is that we replace the max pooling with the average pooling.
As illustrated in Fig.~\ref{fig:NISDenseEdgeConv}, dense connections are employed both within and between spiking graph convolution layers.
In each graph convolution layer, $H_\theta$ is densely connected, and features are passed to all subsequent layers.
These dense connections reduce network parameters and enhance contextual information\cite{liu2019densepoint}.

\subsection{Noise-Injected Spiking Graph Convolutional Networks}
\label{subsec:net}

\begin{figure*}[!t]
    \begin{center}
    \includegraphics[width=1\linewidth]{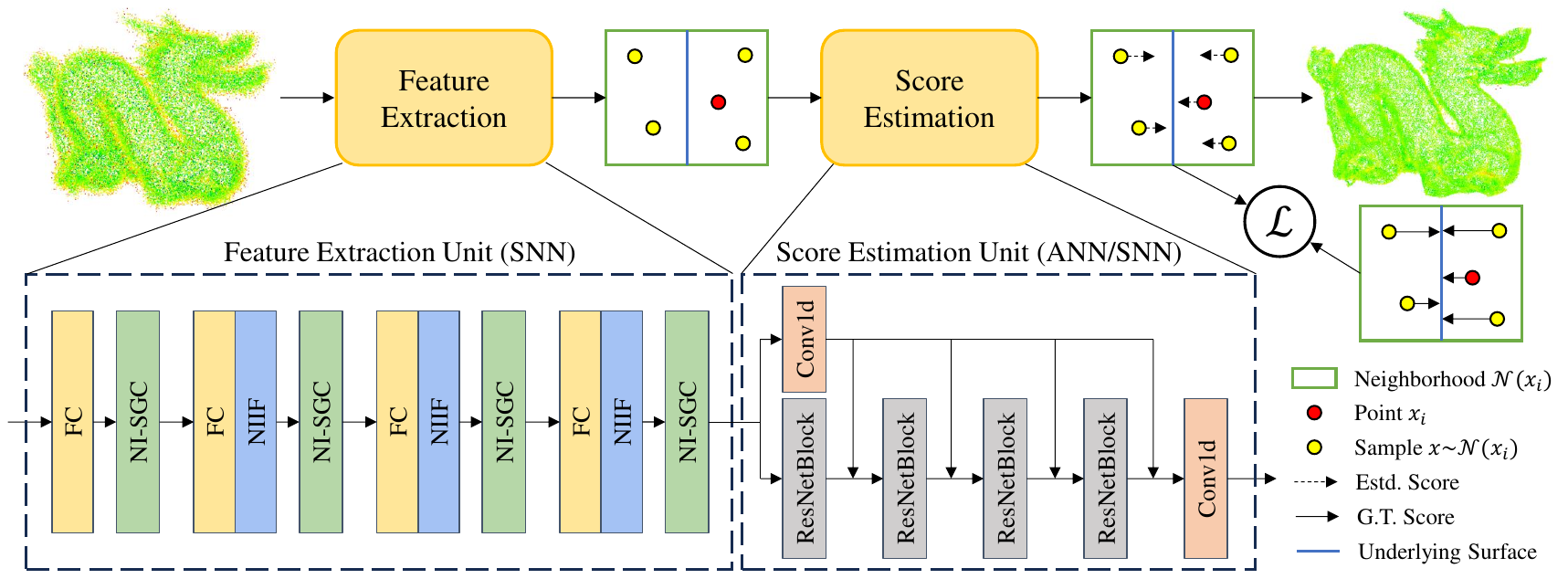}
    \end{center}
    \caption{Illustration of the architecture and pipeline of the noise-injected spiking point cloud denoise learning network.}
    \label{fig:net}
\end{figure*}
In this section, we design noise-injected spiking graph convolutional networks (NI-SGCN) for 3D point cloud denoising, which is based on an ANN-based denoising architecture, ScoreDenoise~\cite{luo2021score}. The NI-SGCN consists of two main parts: the spiking feature extraction module and the score estimation module.

The spiking feature extraction module aims to learn point-wise spiking features from the input noisy 3D point set $X=\left\{x_i\right\}_{i=1}^N \in R^{T*N*3}$, where $T$ represents the spiking time latency, $N$ denotes the number of points, and $3$ represents the 3D coordinates of points. 
The structures of our spiking feature extraction module are shown in Fig.~\ref{fig:net}, 
which is constructed by stacking dense NI-SGC blocks and fully connected layers. 
The learned spiking feature for point $x_i$ is denoted as $h_i$.

The score estimation module is parameterized by the feature $h_i$ of point $x_i$,
which outputs a score $Sc(x_i)$. The $Sc(x_i)$ represents the gradient from $x_i$ to the underlying surface and is used to determine the direction of optimizing noise.
There are two variants: one is the ANN-based score estimation, following ScoreDenoise~\cite{luo2021score} and the other is a SNN-based score estimation by simply replacing the ReLU in the ANN-based score estimation with NIIF spiking neurons.
Consequently, we propose two implementations of NI-SGCN: the first is a hybrid architecture, NI-HSGCN, witch employs the above spiking feature extraction module and the ANN-based score estimation module; 
the second is NI-PSGCN, a purely SNN-based structure, which fully leverages the enhanced energy efficiency of SNNs.

The final training objective aggregates the objectives for each local score function:
\begin{equation}
\mathcal{L}=\frac{1}{N} \sum_{i=1}^N \mathbb{E}_{j \sim \mathcal{N}\left(i\right)}\left[\left\|sc(x_j)-Sc(x_i)\right\|_2^2\right]
\end{equation}
where $\mathcal{N} (i)$ is a distribution concentrated in the neighborhood of $x_i$.
$sc(x_j)$ is the ground truth score.
Note that, this objective not only matches the predicted score on the position of $x_i$ but also matches the score on the neighboring areas of $x_i$. 
For NIIF neuron, we only add noise during training to improve robustness.
The derivative of the Heaviside step function in Eq.~(\ref{eq:update}) equals the Dirac delta function, which makes the training process unstable if used directly for gradient descent.
Following~\cite{fang2023spikingjelly}, we use the surrogate function $\Theta'(v) \triangleq \sigma'(v)$ in gradient back-propagation, where $\sigma(v)$ is a smooth, continuous function resembling $\Theta(v)$, specifically the arctan function.

\section{Experiments and Results}
\label{sec:exp}
\subsection{Experimental Settings}
\label{subsec:setting}
We evaluate the performance of 3D point cloud denoising using two benchmark datasets, PU-Net~\cite{yu2018pu} and PC-Net~\cite{rakotosaona2020pointcleannet}.
PU-Net contains 60 distinct shapes, which are divided into 40 for training and 20 for testing. 
The PC-Net dataset, used exclusively for generalization testing, comprises 10 unique point cloud shapes and their corresponding meshes.

Following the experimental setup in the literature~\cite{luo2021score}, the meshes in both datasets are normalized to a unit sphere.
We use Poisson disk sampling to generate point cloud data at a resolution of 10K and 50K points. Then we add Gaussian noise with a standard deviation of 0.5\% to 2\% of the unit sphere.
The model is trained on the PU-Net training set, and its denoising performance is evaluated on both the PU-Net test set and the PC-Net dataset.
In line with previous work, Chamfer Distance (CD) and Point-to-Mesh Distance (P2M) are used as evaluation metrics, with both CD and P2M reported in units of $10^{-4}$.

All experiments are implemented on Intel an i9-13900HX CPU and an NVIDIA RTX 4090 GPU (24GB memory, CUDA 11.8), using PyTorch and SpikingJelly~\cite{fang2023spikingjelly} for implementation. 
NI-SGCN is trained with the Adam optimizer using a learning rate of $1\times10^{-4}$, and the network is trained with a batch size of 32. 
For all our SNN models, the default time delay $T$ is set to 4, the membrane potential threshold is set to 1, and the standard deviation of injected noise is 0.2.

\subsection{Ablation Study}
\label{subsec:abl}
\begin{table}[!t]
    \centering
    \setlength{\tabcolsep}{10pt}
    \begin{tabular}{c|cc}
        \hline
        module      & CD                & P2M\\
        \hline
        NI-HSGCN    & \textbf{1.798}    &\textbf{1.147}   \\
        NI-PSGCN    & 1.926             & 1.211  \\
        \hline
    \end{tabular}
    \caption{Ablation study on the NI-SGCN architecture designs on the PU-Net dateset.}
    \label{tab:score}
\end{table}

We first perform ablation experiments on PU-Net to establish the final architecture of NI-SGCN.
All ablation study are performed on the PU-Net dataset, using point clouds with 50K points and 2\% Gaussian noise.

\paragraph{Ablation on NI-SGCN architecture designs.}
Starting from the spiking graph convolution, we build two SNN-based denoising networks.
One is a pure spike graph convolutional network structure NI-PSGCN, and the other is a hybrid architecture NI-HSGCN, which integrates some ANN-based learning operations. 
We use PU-Net to perform denoising ablation experiments on both architectures, with the evaluation results reported in Table~\ref{tab:score}.
The hybrid NI-HSGCN outperforms the purely spiking NI-PSGCN, with a reduction of 0.128 in CD and a reduction of 0.064 in P2M metrics.
While NI-PSGCN has achieved good denoising results using a pure SNN, its regression module (score estimation) cannot match the structural performance of an ANN. However, by adopting a hybrid structure, NI-PSGCN leverages the complementary strengths of both SNN and ANN, balancing energy efficiency with denoising effectiveness. 
Therefore, we choose the hybrid NI-HSGCN architecture to build the final denoising network in subsequent experiments.

\paragraph{Ablation on pooling schemes.}
\begin{table}[!t]
    \centering
    \setlength{\tabcolsep}{13pt}
    \begin{tabular}{c|cc}
        \hline
        Decoding Scheme&   CD   & P2M\\\cline{1-3}
        \hline
        MAX        & 2.199    &1.495   \\
        MEAN    & \textbf{1.798} & \textbf{1.147} \\
        \hline
    \end{tabular}
    \caption{Ablation study on the pooling schemes on the PU-Net dataset. }
    \label{tab:pooling schemes}
\end{table}
We perform a series of ablation studies comparing max pooling and mean pooling, with the results shown in Table~\ref{tab:pooling schemes}. 
The findings show that the NI-SGCN network achieves superior denoising performance with mean pooling.
Specifically, mean pooling reduces the CD by $0.401$ and the P2M by $0.348$ compared to max pooling. In both metrics, mean pooling consistently outperforms max pooling.
This improvement is attributable to the binary outputs (0 or 1) produced by spiking neurons in SNNs, in contrast to the float values generated by activation functions in ANNs. 
Max pooling in SNNs captures only the most prominent binary features, potentially overlooking finer details. 
In contrast, mean pooling converts binary features into float values, preserving more information about the underlying geometry. 
Consequently, mean pooling offers a more representative and stable aggregation of local geometric features.

\paragraph{Ablation on spiking neurons.} 
\begin{table}[!t]
    \centering
    \setlength{\tabcolsep}{5pt}

    \begin{tabular}{c|ccccccc}
        \hline
        neurons & IF & LIF & NIIF & NILIF & QIF & KLIF\\
        \hline
        CD & 2.057 & 1.889 & \textbf{1.798} & 1.848 & 2.176 & 2.469 \\
        P2M & 1.362 & 1.203 & \textbf{1.147} & 1.186 & 1.465 & 1.711 \\
        \hline
    \end{tabular}
    \caption{Ablation study on the usage of different spiking neurons in the network.}
    \label{table:neurons_comparison}
\end{table}
We assess the impact of various types of spiking neurons on the performance of NI-SGCN. Specifically, we compare the network's performance using NIIF, Noise-injected Leaky Integrate-and-Fire (NILIF), IF, LIF, QIF~\cite{brunel2003firing} and KLIF~\cite{jiang2023klif} neurons. 
As shown in Table~\ref{table:neurons_comparison}, NIIF neuron achieve the best performance in both CD and P2M metrics.
Noise-injected IF neurons can directly incorporate noise into the integration process, thereby enhancing the robustness and generalization ability of the network.
Spiking neurons with noise-perturbed dynamics are believed to be more biologically realistic, and internal noise brings potential benefits by promoting more generalization performance.
Furthermore, the simplicity of the IF neuron, when combined with noise injection, allows for more efficient and effective spike-based computation compared to the LIF neurons.

\paragraph{Ablation on time latency.} 
\begin{table}[!t]
    \centering
    \setlength{\tabcolsep}{10pt}
    \begin{tabular}{c|cccc}
    \hline
    Latency &   1       &  2        &  4                    &  8\\
    \hline
    CD      & 5.524     & 1.951     & \textbf{1.798}        & 1.815\\
    P2M     & 4.419     & 1.273     & \textbf{1.147}        & 1.155\\
    \hline
    \end{tabular}
    \caption{Ablation study on the time latency on the PU-Net dataset. NI-SGCN with $T=4$ presents the highest CD and P2M metrics.}
    \label{tab:latency}
\end{table}
In spiking neural networks, the time delay $T$ is a critical hyperparameter. As illustrated in Table~\ref{tab:latency}, our evaluation indicates that the network sustains strong performance metrics with minimal variation in denoising effectiveness when $T$ ranges from $2$ to $8$. Notably, the optimal denoising performance, characterized by the lowest CD and P2M metrics, occurs at $T=4$.
While increasing $T$ from $2$ to $4$ improves denoising performance, further increasing $T$ to $8$ results in a decline, likely because excessively large $T$ values introduce redundancy rather than providing valuable information.

\subsection{Comparison with State-of-the-art Methods}
\label{subsec:com}

\begin{table*}[!t]
    \centering
    \setlength{\tabcolsep}{10pt}

    \begin{tabular}{c|c|l|cc|cc|cc|cc}
        \hline
         \multicolumn{3}{c|}{Points}  & \multicolumn{4}{c|}{10K (Sparse)} & \multicolumn{4}{c}{50K (Dense)}\\
        \hline
         \multicolumn{3}{c|}{Noise}  & \multicolumn{2}{c|}{1\%} & \multicolumn{2}{c|}{2\%}& \multicolumn{2}{c|}{1\%} & \multicolumn{2}{c}{2\%} \\
        \hline
         Dataset & \multicolumn{2}{c|}{Model}  & CD & P2M & CD & P2M & CD & P2M & CD & P2M \\
        \hline
          \multirow{5}*{PU} & \multirow{4}*{ANN} & PCN & 3.686 & 1.599 & 7.926 & 4.759 & 1.103 & 0.646 & 1.978 & 1.370 \\
          ~ & ~ & ScoreDenoise & \textbf{2.611}  & \textbf{0.863} & \textbf{3.684} & \textbf{1.416} & 0.767 & 0.448 & 1.295 & 0.842 \\
          ~ & ~ & DMRDenoise & 4.712 & 2.196 & 5.085 & 2.523 & 1.205 & 0.762 & 1.443 & 0.970 \\
          ~ & ~ & Pointfilter & 2.709 & 0.884 & 4.508 & 1.937 & \textbf{0.723} & \textbf{0.389} & \textbf{1.175} & \textbf{0.709} \\
          \cline{2-11}
          ~ & \multirow{2}*{SNN} & NI-HSGCN & 2.797 & 0.923 & 4.437 & 1.884 & 0.843 & 0.461 & 1.798 & 1.147 \\
          ~ & ~ & NI-PSGCN & 2.947 & 1.004 & 4.622 & 1.976 & 0.885 & 0.485 & 1.926 & 1.211 \\
        \hline
        \multirow{5}*{PC} & \multirow{4}*{ANN} & PCN & 3.847 & \textbf{1.221} & 8.752 & 3.043 & 1.293 & \textbf{0.289} & 1.913 & 0.505 \\
          ~ & ~ & ScoreDenoise & \textbf{3.264} & 1.663 & \textbf{5.066} & 2.485 & 1.075 & 0.543 & 1.671 & 1.006 \\
          ~ & ~ & DMRDenoise & 6.602 & 2.152 & 7.145 & \textbf{2.237} & 1.566 & 0.350 & 2.009 & \textbf{0.485} \\
          ~ & ~ & Pointfilter & 3.374 & 1.945 & 6.160 & 3.480 & \textbf{1.060} & 0.522 & \textbf{1.620} & 0.982 \\
          \cline{2-11}
          ~ & \multirow{2}*{SNN} & NI-HSGCN & 3.478 & 1.799 & 5.756 & 3.073 & 1.146 & 0.571 & 2.134 & 1.357 \\
          ~ & ~ & NI-PSGCN & 3.798 & 1.972 & 6.035 & 3.380 & 1.208 & 0.626 & 2.258 & 1.441 \\
        \hline
    \end{tabular}
    \caption{Comparison among competitive denoising algorithms. The units of CD and P2M are both $10^{-4}$.}
    \label{tab:sota}
\end{table*}

\begin{figure*}[!t]
    \centering
    \includegraphics[width=0.95\linewidth]{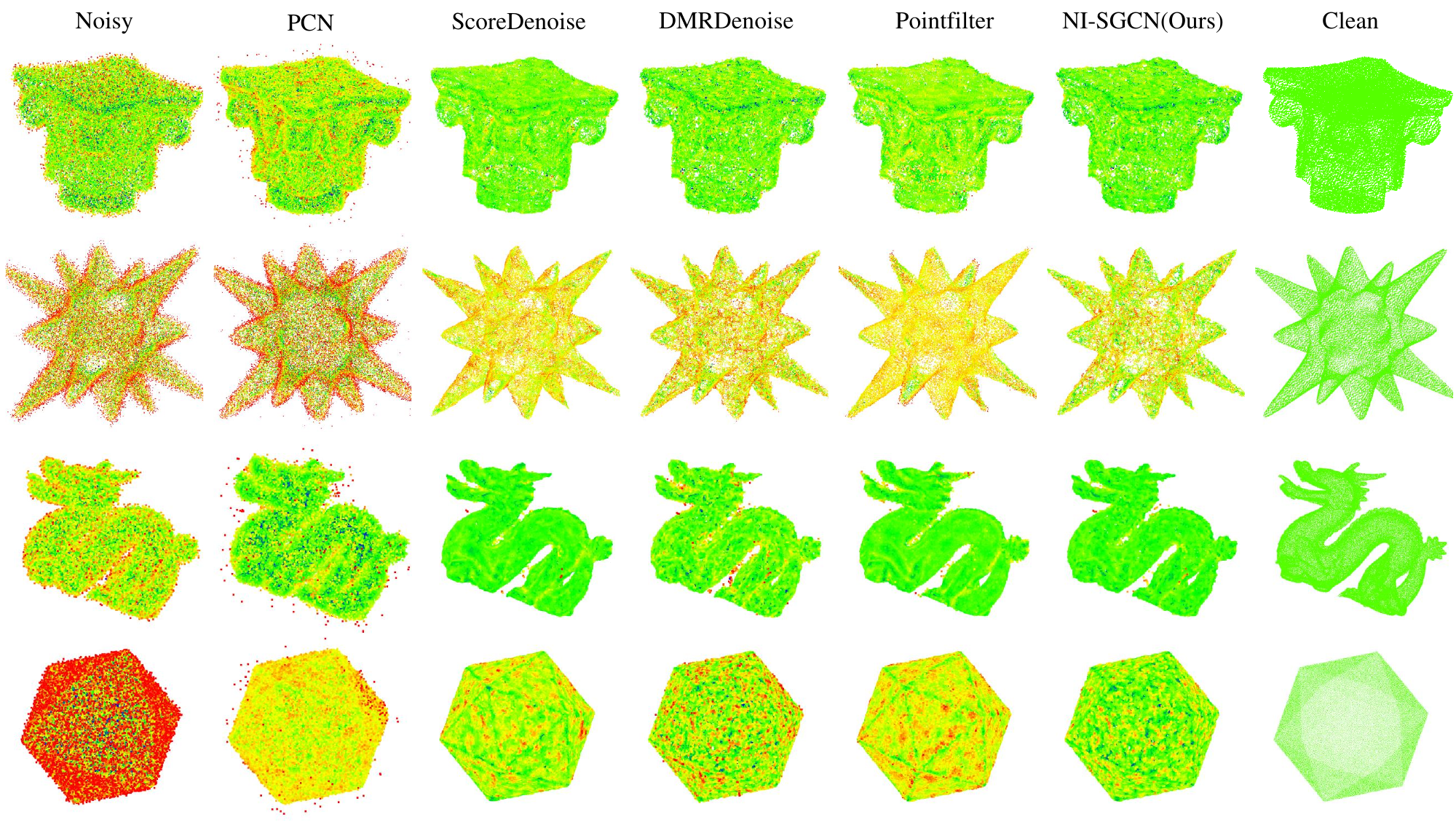}
    \caption{Visual comparison of denoising methods. Points colored reder are farther away from the ground truth surface.}
    \label{fig:sota-pic}
\end{figure*}

We compare our method against state-of-the-art deep learning-based denoisers, including PCN~\cite{rakotosaona2020pointcleannet}, ScoreDenoise~\cite{luo2021score}, DMRDenoise~\cite{luo2020differentiable}, and Pointfilter~\cite{zhang2020pointfilter}. Our evaluations are conducted on the PU-Net and PC-Net datasets, with isotropic Gaussian noise applied at a standard deviation ranging from 1\% to 2\% of the unit sphere radius. 

As shown in Table~\ref{tab:sota}, the proposed NI-HSGCN and NI-PSGCN maintain low accuracy loss 
compared with the ANN-based ScoreDenoise and Pointfilter. However, our work has the advantage
of ultra-low energy consumption as demonstrated in the next section.
These results indicate the capability of NI-HSGCN and NI-PSGCN in preserving the point cloud structure while effectively reducing noise.
Our work can provide a proper balance between denoising accuracy and efficiency.
On the other hand, our networks outperform PCN and DMRDenoise in most cases. 
The experiments across different datasets and noise levels further highlight the robustness and adaptability of our networks.

Figure~\ref{fig:sota-pic} visually compares the denoising results achieved by our proposed method with those of competitive baselines, under $2\%$ Gaussian noise and a point cloud resolution of $50K$ points. Each point is color-coded according to its denoising error, as measured by the point-to-grid distance, with green indicating points closer to the underlying surface and red indicating points farther away.
The visualization, consistent with Table~\ref{tab:sota}, clearly demonstrates that our NI-HSGCN produces cleaner and more visually appealing outcomes compared to PCN and DMRDenoise. 
Our work can reduce noise and preserve fine details to some extents.

\subsection{Theoretical Energy Consumption Calculation}
\label{sec:energy}
We examine the hardware efficiency of the proposed framework.
In contrast, our SNN architecture leverages a transformation that largely bypasses multiplication, retaining it only in the initial layer.
This design allows the hardware to take advantage of sparse computation, effectively eliminating addition operations in the absence of spikes.
We estimate the primary energy consumption of our network and the ScoreDenoise~\cite{luo2021score} network, while excluding point cloud downsampling and normalization operations.
Both networks are tested with an input of $5,000$ 3D points.
Following~\cite{horowitz20141}, by quantifying the computational workload in terms of operations executed, the conventional ANN-based model theoretically consumes $3.01 \times 10^9pJ$ for per forward pass,
In contrast, the SNN-based architecture NI-PSGCN requires only $2.36 \times 10^8pJ$, representing a $12.75$-fold reduction in energy consumption.
The hybrid architecture NI-HSGCN consumes $7.65 \times 10^8pJ$, which is approximately $3.94$ times more energy-efficient than the traditional ANN.
Due to the sparsity of spikes and the use of alternating current, our network demonstrates exceptional energy efficiency.
Our work underscores the potential of SNNs for 3D point cloud denoising and supports the development of energy-efficient 3D data acquisition devices.

\section{Conclusion}
In this paper, we propose noise-injected spiking graph convolutional networks for 3D point cloud denoising, achieving an optimal balance between denoising effectiveness and bio-inspired energy efficiency on two benchmark datasets, PU-Net and PC-Net. In future work, we aim to explore more efficient and energy-saving 3D point cloud denoising networks while maintaining the high accuracy demonstrated by state-of-the-art methods.

\section{Acknowledgments}
This work was supported by the National Natural Science Foundation of China (No.52275493, No.92367301, No.92267201, No.92160301, No.52425506, No.62206001).

\bibliography{aaai25}

\end{document}